\documentclass[9pt]{article} 
\usepackage{spconf,amsmath,graphicx,amssymb}

\usepackage{caption}
\usepackage{subcaption}
\usepackage{algorithm}
\usepackage{algpseudocode}
\usepackage{bbm}
\usepackage{graphicx}

\def\balph{{\boldsymbol{\alpha}}}
\newcommand{\argmin}{\operatornamewithlimits{argmin}}

\algnewcommand\algorithmicinput{\textbf{Input:}}
\algnewcommand\INPUT{\item[\algorithmicinput]}
\algnewcommand\algorithmicoutput{\textbf{Output:}}
\algnewcommand\OUTPUT{\item[\algorithmicoutput]}

\title{A multi-layer image representation using Regularized Residual Quantization: application to compression and denoising
}
\name{Sohrab Ferdowsi, Slava Voloshynovskiy, Dimche Kostadinov}
\address{Department of Computer Science, University of Geneva, Switzerland\\
$\lbrace$sohrab.ferdowsi, svolos, dimche.kostadinov$\rbrace$@unige.ch
}
\begin{document}
%
\maketitle
\begin{abstract}
A learning-based framework for representation of domain-specific images is proposed where joint compression and denoising can be done using a VQ-based multi-layer network. While it learns to compress the images from a training set, the compression performance is very well generalized on images from a test set. Moreover, when fed with noisy versions of the test set, since it has priors from clean images, the network also efficiently denoises the test images during the reconstruction. The proposed framework is a regularized version of the Residual Quantization (RQ) where at each stage, the quantization error from the previous stage is further quantized. Instead of codebook learning from the k-means which over-trains for high-dimensional vectors, we show that only generating the codewords from a random, but properly regularized distribution suffices to compress the images globally and without the need to resort to patch-based division of images. The experiments are done on the \textit{CroppedYale-B} set of facial images and the method is compared with the JPEG-2000 codec for compression and BM3D for denoising, showing promising results. 

\end{abstract}
\section{Introduction}
\label{sec:intro}
Consider the classical image processing tasks like image compression and denoising. While there exists a wealth of successful methods to address them, the specificity and the intricate optimization in their design hinders their application to more general tasks and setups. For example suppose instead of one single image, we are given a collection of similar-looking images. Can the standard image compression codecs benefit from the shared redundancy to compress the images further? Such a setup is of great practical importance for compression of facial or iris images in biometrics, medical images, or the compression and transmission of very large, but similar-looking images in remote sensing and astronomy. In these cases, the usage of generic codecs like JPEG-2000 whose basis vectors are not adapted to the statistics of images is known to be inefficient.

Take the case of facial images. Inspite of the extensive litereture in generic image compression, only several learning-based algorithms have studied the compression of facial images. For example, \cite{4286990} was an early attempt based on VQ. \cite{Bryt2008270} learns the dictionaries based on the K-SVD \cite{1710377} while \cite{6844846} uses a tree-based wavelet transform. \cite{5740941} proposes a codec by using the Iteration Tuned and Aligned Dictionary (ITAD). In spite of their high compression performance, the problem with most of these approaches is that they rely very much on the alignment of images and they are less likely to generalize once the imaging setup is changed a bit. Some of them require the detection of facial features (sometimes manually) and then alignment by geometrical transformation into some canonical form and also a background removal stage. 

Similarly for the image denoising tasks, only few methods have benefited from external clean databases of similar images. For example, \cite{7063913} reports a near $1dB$ improvement over the BM3D.

On the other hand, one can think of different tasks to be performed jointly. Can more favorable scenarios like the availability of a collection of similar domain-specific images help to compress and denoise images at the same time? As a practical scenario, suppose for example the case where in an object identification system, several exemplar images have been taken with high-quality acquisition systems in the enrollment mode. At query time, however, only low-quality and noisy cameras are available. It is highly desirable to be able to jointly denoise and compress the acquisitions.

The rest of the paper is organized as follows. In section \ref{SOTA}, a very brief overview of the general image representation formulation is considered where several relevant cases are quickly reviewed. Section \ref{proposed} preludes with a review from a problem in rate-distortion theory, namely the reverse water-filling paradigm. This will be used as the core concept behind the proposed Regularized Residual Quantization (RRQ) introduced next. Section \ref{exp} conducts experiments on the RRQ algorithm under the image compression and denoising tasks. Finally, section \ref{summary} concludes the paper.

\section{Related work}
\label{SOTA}

Many methods for image representation and dictionary learning can be generalized in the inverse-problem formulation of Eq. \ref{eq:general_quantization}, where $\mathrm{X}$ contains the data-points (e.g., image patches) $\mathbf{x}(i)$'s in its columns. The codebook and the codes can be represented in a matrix form as $\mathrm{C} = [\mathbf{c}_1,\cdots, \mathbf{c}_k, \cdots,\mathbf{c}_K]$\footnote{Notation: matrix $\mathrm{X}$, random variable $X$, random vector $\mathbf{X}$ and vector $\mathbf{x}$} and $\mathrm{A} = [\balph_1, \cdots, \balph_i, \cdots, \balph_N]$, respectively, with $1 \leqslant i \leqslant N$ and $1 \leqslant k \leqslant K$. 

\begin{equation} \label{eq:general_quantization}
\begin{aligned}
& \underset{{\mathrm{C},A}}{\text{minimize}}
& & ||\mathrm{X} - \mathrm{C} \mathrm{A} ||_F^2. \\
& \text{s.t.}
& & \Omega_{\mathrm{C}}, \Omega_{\mathrm{A}}  \\
\end{aligned}
\end{equation}
$\Omega_{\mathrm{C}}$ and $\Omega_{\mathrm{A}} $ are a set of constraints on the construction of the codebook and the codes, respectively.

Depending on $\Omega_{\mathrm{C}}$ and $\Omega_{\mathrm{A}} $, the problem of Eq. \ref{eq:general_quantization} can be treated in many different ways.\footnote{See \cite{5452966} and \cite{CGV-058} for detailed reviews and discussions} For example, under the famous sparsity constraint $\Omega^{\ell_0}_{\mathrm{A}}: ||\balph(i)||_0 \leqslant s$ or its relaxed version $\Omega^{\ell_1}_{\mathrm{A}}: ||\balph(i)||_1 \leqslant s$, the K-SVD algorithm  \cite{1710377} solves it for local minima in an iterative way. 

In this work, we follow the VQ-based interpretation of Eq. \ref{eq:general_quantization}, where, as a general formulation, it is required that:

\begin{equation*}
\Omega_{\mathrm{A}} ^{\text{VQ}}: ||\balph_i||_0 = ||\balph_i||_1 = 1. \label{eq:VQ_code_cstr}
\end{equation*}

This problem can be solved using the k-means algorithm. However, the lack of structure for this formulation leads to poor generalization performance. To address some of the issues with this simple formulation, Product Quantization (PQ) (e.g., \cite{Gersho:1991:VQS:128857} \cite{5432202}) divides the vectors into several blocks and runs k-means on each of them independently. While PQ can achieve good rate-distortion performance under certain conditions, its lack of design flexibility and the fact that the system should be re-trained for every rate, makes PQ not a suitable solution for image analysis.   
  
 As an alternative, RQ is a multi-layer approach that at each layer quantizes the residuals of quantization of the previous layer. While having been extensively studied in the 80's and 90's for different tasks like image coding (e.g., refer to \cite{480761}, \cite{3776} or \cite{Gersho:1991:VQS:128857}), its efficiency was limited for more modern applications. In practice, it was not possible to learn codewords for more than a couple of layers.

In this work, we use an approach based on the RQ for which we introduce a pre-processing and an efficient regularization, making it possible to learn arbitrary numbers of layers. Moreover, the introduced regularization makes it possible to go beyond the image patches and work with the high-dimensional image directly. This brings an important advantage for different tasks like image compression. Since the global picture of the image is preserved in the high-dimensional representation, one does not have to encode the relation between similar patches after compression.

\section{Proposed framework: RRQ} \label{proposed}
We first recall a concept from rate-distortion theory which is the quantization of Gaussian independent sources. Although in a slightly different setup than a practical quantization (e.g. being asymptotic), this motivates the core idea behind the RRQ algorithm introduced next in this chapter.

\subsection{Preliminaries: Quantization of independent sources} \label{sub:waterfilling}
The trade-off between the compactness and the fidelity of representation of a signal is classically treated in the Shannon's rate-distortion theory \cite{shannoncoding59}\footnote{Refer to Ch. 10 of \cite{CoverThomas200607} for further details of this subsection.}.

A special setup studied in this theory is the rate-distortion for $n$ independent Gaussian distributed sources, $X_j$'s with different variances. Concretely, assume $X_j \sim \mathcal{N}(0,\sigma_j^2)$. Define the expected distortion between a random vector and its estimate as $\mathcal{D} \triangleq \mathbb{E}[d(\mathbf{X},\hat{\mathbf{X}})]$, where the distortion between two n-vectors $\mathbf{a}$ and $\mathbf{b}$ is defined as $d(\mathbf{a},\mathbf{b}) \triangleq \frac{1}{n} ||\mathbf{a} - \mathbf{b}||_2^2$.

Here we ask the question: Given a fixed total distortion $D$ allowed, i.e., $\mathcal{D} \leqslant D$, what is the optimal way to divide the distortion (or rate) between these sources such that the overall allocated rate (distortion) is minimized? This can be posed as:
\begin{equation} \label{eq:water-fill}
\begin{aligned}
& \underset{D_j}{\text{min}}  &  \sum_{j = 1}^n \text{max}\big[0,\frac{1}{2} \log_2 \frac{\sigma_j^2}{D_j}\big], \\
& \text{s.t.}  & \sum_{j = 1}^n D_j = D, \\
\end{aligned}
\end{equation}
where $D_j$ is the distortion of each source after rate-allocation. The solution to this convex problem is known as the reverse water-filling and is given as:

\begin{equation} \label{eq:water-fill_distortion}
D_j =
\begin{cases}
   \gamma ,& \text{if   } \sigma_j^2 \geqslant \gamma \\
    \sigma_j^2, & \text{if   } \sigma_j^2 < \gamma.
\end{cases}
\end{equation} 
where $\gamma$ is a constant which should be chosen to guarantee that $\sum_{j=1}^n D_j = D$.

Denote $\sigma_{C_j}^2$, the variance of the codewords for quantization of $X_j$. Due to the principle of orthogonality and the independence of dimensions, we have that $\sigma_{C_j}^2 = \sigma_j^2 - D_j$. Therefore, according to Eq. \ref{eq:water-fill_distortion}, the optimal assignment of the codeword variances will be a soft-thresholding of $\sigma_j^2$ with $\gamma$:
\begin{equation} \label{eq:Sig2C_waterfill}
\sigma_{C_j}^2 = \Big( \sigma_j^2 - \gamma \Big)^+ = 
\begin{cases}
   \sigma_j^2 - \gamma ,& \text{if   } \sigma_j^2 \geqslant \gamma \\
    0, & \text{if   } \sigma_j^2 < \gamma.
\end{cases}
\end{equation} 

This means that the optimal rate-allocation requires that the sources with variances less than $\gamma$ should not be assigned any rate at all. This, when used in the codebook design, results in sparsity of the codewords which we incorporate in the RRQ algorithm.

\begin{algorithm} \caption{Pre-processing} \label{alg:preprocessing}
\begin{algorithmic}[0]
    \INPUT  $\mathcal{I}_{\text{train}} = \lbrace \mathrm{I}(i), \cdots, \mathrm{I}(N)\rbrace$, images in the train set.
    \OUTPUT $\mathrm{X}$, matrix of decorrelated vectors, $\mathrm{V}_1, \cdots, \mathrm{V}_M$, rotation matrices for the sub-bands. 
\end{algorithmic}
\begin{algorithmic}[1]
\For{$i \in  \mathcal{I_{\text{train}}}$}
\State $\mathrm{I}'(i) \gets \text{2D-DCT}[\mathrm{I}(i)]$
\State $\mathbf{x}'(i) \gets \text{zig-zag}[\mathrm{I}'(i)]$ \Comment: zig-zag vectorization
\State Divide $\mathbf{x}'(i)$ into $M$ equal sub-bands: $\mathbf{x}'(i) = [\mathbf{x}_1'(i), \cdots, \mathbf{x}_M'(i)]$  
\For{$m = 1, \cdots, M$}
\State Stack all $\mathbf{x}_m'(i)$'s to get $\mathrm{X}_m $
\EndFor
\EndFor
\For{$m = 1, \cdots, M$}
\State Perform PCA on $\mathrm{X}'_m$ (without dim. reduction) to get $\mathrm{X}_m$ and $\mathrm{V}_m$, the rotation matrix
\State Concatenate $\mathrm{X}_m$'s to get $\mathrm{X}$ 
\EndFor

\end{algorithmic}
\end{algorithm}


\begin{algorithm} \caption{Regularized Residual Quantization} \label{alg:RRQ}
\begin{algorithmic}[0]
    \INPUT  de-correlated train set $\mathrm{X}$ 
    \OUTPUT multi-layer codebooks $\mathrm{C}^{(l)}$'s and index sets $\mathrm{A}^{(l)}$'s, with $l=1,\cdots,L$
\end{algorithmic}
\begin{algorithmic}[1]
\State $\hat{\mathrm{X}} \gets \mathrm{O}$
\State $\mathrm{X}^{(0)} \gets \mathrm{X}$
\State $[D_1^{(0)}, \cdots, D_n^{(0)}] \gets \text{var}[\mathrm{X}^{(0)}]$ \Comment variance per dimension
\For{$l = 1, \cdots, L$} 
\State $\gamma^* \gets \underset{\gamma}{\text{argmin}} \big(|\log_2{K^{(l)}} - {\underset{j \in \mathcal{A}_{\gamma}}{\sum} \frac{1}{2}\log_2{\frac{D_j^{(l-1)}}{\gamma}}}|  \big)$
\For{$j = 1, \cdots, n$}
\State  $\sigma_{C_j^{(l)}}^2 \gets \Big( D_j^{(l-1)} - \gamma^* \Big)^+$
\EndFor
\State $\mathrm{S}^{(l)} \gets \text{diag}(\sigma_{C_1^{(l)}}^2, \cdots, \sigma_{C_n^{(l)}}^2)$
\For{$k = 1, \cdots, K^{(l)}$}
\State $\mathbf{c}_k^{(l)} \gets$ Generate randomly from $\mathcal{N}(\mathbf{0},\mathrm{S}^{(l)})$
\State Concatenate $\mathbf{c}_k^{(l)}$'s to get $ \mathrm{C}^{(l)}$
\EndFor

\For{$i = 1, \cdots,N$}
\State $k^* \gets \argmin_{1 \leqslant k \leqslant K^{(l)}} || \mathbf{x}(i)^{(l-1)} - \mathbf{c}_k^{(l)}||_2 $
\State $\balph(i)^{(i)} \gets $ all-zero vector with $1$ at the $k^*$ position
\State Concatenate $\balph(i)^{(l)}$'s to get $\mathrm{A}^{(l)}$
\EndFor
\State $\hat{\mathrm{X}}^{(l)} \gets \mathrm{C}^{(l)} \mathrm{A}^{(l)}$
\State $\hat{\mathrm{X}} \gets \hat{\mathrm{X}} + \mathrm{X}^{(l)}$
\State $\mathrm{X}^{(l)} \gets  \mathrm{X}^{(l-1)} - \hat{\mathrm{X}}^{(l)}$
\State $[D_1^{(l)}, \cdots, D_n^{(l)}] \gets \text{var}[\mathrm{X}^{(l)}]$
\EndFor

\end{algorithmic}
\end{algorithm}


\subsection{The RRQ algorithm} \label{sub:RRQ}
Inspired by the setup studied in section \ref{sub:waterfilling}, we argue that after a pre-processing stage, natural images can be represented in a global representation as variance decaying vectors which have independent, or at least uncorrelated dimensions. One might think of the PCA as a simple way to achieve this. However, since the dimensionality of the entire vectorized image is high, apart from the big complexities incurred, there will be too many parameters in the covariance matrix to estimate. Therefore, a global PCA will likely over-fit to the training, largely deviating from the test set. To overcome this issue, we propose the pre-processing in Algorithm \ref{alg:preprocessing}.

After the PCA rotation matrices are learned from the training set, the same procedure applies to images from the test set. In fact, this pre-processing is a more robust estimation for the global PCA. Instead of $n^2$ parameters of the direct PCA rotation matrix, with the help of 2D-DCT, this pre-processing has $m (\frac{n}{m})^2$ parameters to estimate. This is an effective way to trade independence of dimensions for robustness between train and test sets. 

The RRQ framework is introduced in Algorithm \ref{alg:RRQ}. For each of the $L$ layers, given the desired number of codewords, $K^{(l)}$, after calculation of the variances of the residuals, the algorithm first finds the optimal $\gamma^*$ and calculates the optimal variances of the codewords based on Eq. \ref{eq:Sig2C_waterfill} and then randomly generates $K^{(l)}$ codewords based on these variances. Especially at the first layers, since the data has a very strong decaying character, this makes the codewords very sparse, significantly reducing the complexity and storage cost of the codebooks. The algorithm continues by quantizing the residuals $\mathbf{x}(i)^{(l-1)}$'s with the generated codewords and updating the estimations $\hat{\mathbf{x}}(i)^{(l)}$'s and the new residuals $\mathbf{x}(i)^{(l)}$'s and finishes at the desired $L$.

 \begin{figure*}  
   \begin{center} 
\subcaptionbox{D-R curve (normalized, log-scale)\label{subfig:DR}} {\includegraphics[width=0.33\textwidth]{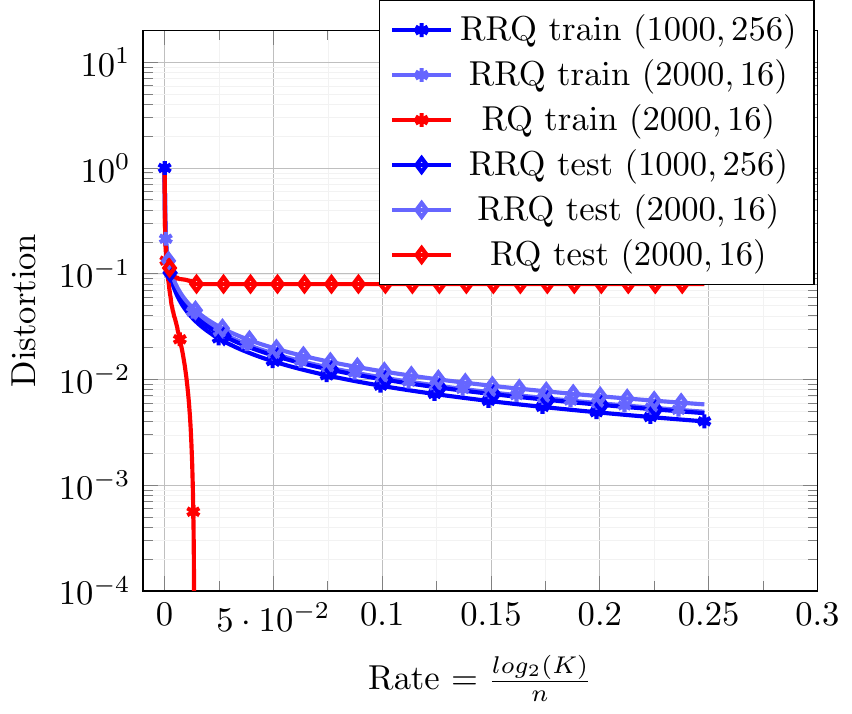}} 
\subcaptionbox{Image compression \label{subfig:comp_CYale}} {\includegraphics[width=0.33\textwidth]{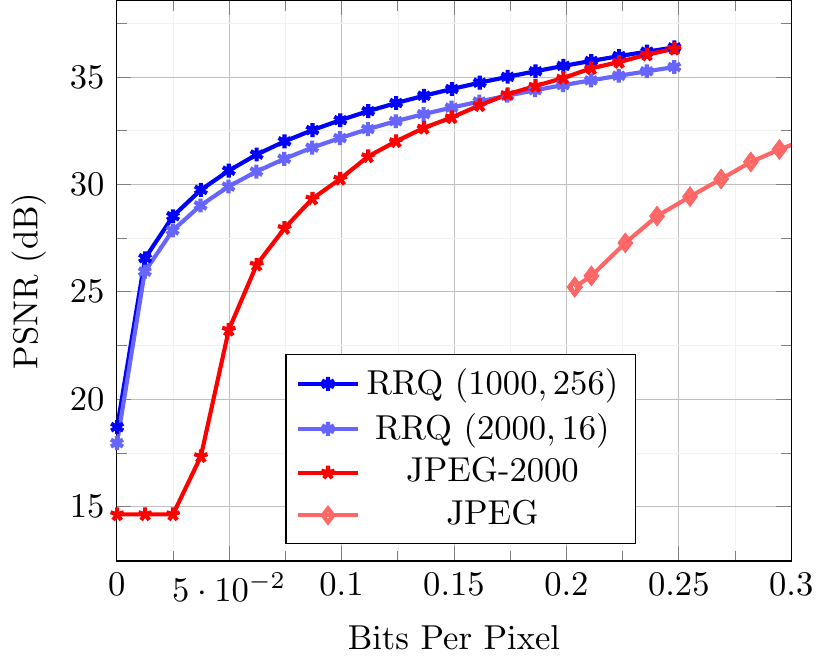}}
\subcaptionbox{Image denoising\label{subfig:Denoising}} {\includegraphics[width=0.33\textwidth]{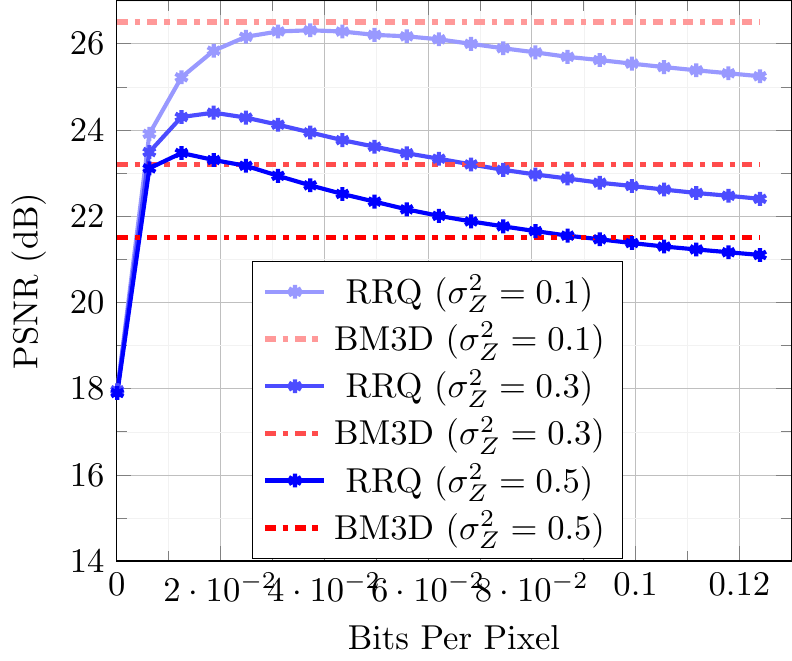}}

   \end{center}
\vspace{-0.5cm}    
   \caption{D-R, compression and denoising average performances on the \textit{CroppedYale-B} set.}
   \label{fig:Results}
   \end{figure*}

\section{Experiments}
\label{exp}
We perform the two tasks of image compression and denoising of facial images. For image compression, we compare the performance of our proposed method with the JPEG and JPEG-2000 standards. For denoising, we compare with the BM3D. These are widely considered as baselines for comparison in the literature. 

The \textit{CroppedYale-B} set \cite{GeBeKr01} is used which contains 2408 images of size $192 \times 168$ from 38 subjects. Each subject has between 57 to 64 acquisitions with extreme illumination changes. We choose half of the images for each subject randomly for training and the rest for testing. 


We choose two different $(L,K)$ value-pairs, $(1000,256)$ and $(2000,16)$, where $L$ is the number of layers and $K$ is the number of codewords per each layer. As described earlier, all codewords are generated randomly according to Eq. \ref{eq:Sig2C_waterfill}. Algorithm \ref{alg:preprocessing} is used for pre-processing with $m = 96$ sub-bands. The resulting decorrelated vectors are of size $n = 192 \times 168 =32256$ (same as the original images).

Figure \ref{subfig:DR} sketches the D-R curve for this set. It is seen that the gap between the training and the test sets for the proposed  RRQ is very small, indicating the success of the algorithm in terms of generalization. The non-regularized RQ, on the other hand, while has much lower distortion on the train set, fails to compress the test set at the first several layers.

%

Fig. \ref{subfig:comp_CYale} shows the results of image compression. These results are averaged over 20 randomly chosen images from the test set. The advantage of the proposed method under this setup over the highly-optimized JPEG-2000 codec is significant, particularly at lower rates. It should be noted that we do not perform any entropy coding over the codebook indices. Further compression improvement can be achieved by entropy coding over the tree-like structure of the codebooks.

The results of image denoising for three different noise levels\footnote{Gray values are normalized between 0 and 1.}, averaged over 20 randomly chosen test images are depicted in Fig. \ref{subfig:Denoising}. The network is trained over clean images and is exactly the same as the one used for compression. Test images are contaminated with noise and are given as the input to the network for reconstruction. When reconstructing the noisy image, the network uses the priors from the clean images based on which it has been trained. These priors are automatically used in the reconstruction process, serving as a very efficient denoising strategy surpassing (only at highly noisy regimes), the prior-less BM3D.

%
As the network tries to reconstruct the noisy image with further details, the noise statistics are becoming more present in the reconstructed image, hence degrading the quality. Therefore, depending on the noise variance, the maximum PSNR is somewhere in the middle of the distortion-rate curve. Noisier images have the maximum at lower rates. 

\begin{figure}  
   \begin{center}
 \captionsetup[subfigure]{labelformat=empty}
   \subcaptionbox{\label{subfig:a}} {\includegraphics[width=0.115\textwidth]{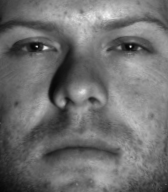}}
    \subcaptionbox{ $\sigma_Z^2 = 0.3$ \label{subfig:b}} {\includegraphics[width=0.115\textwidth]{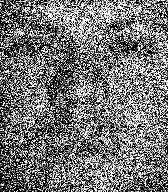}}
    \subcaptionbox{$23.49$ dB\label{subfig:c}} {\includegraphics[width=0.115\textwidth]{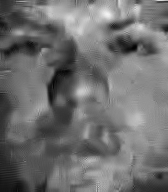}}
   \subcaptionbox{$24.73$ dB\label{subfig:d}} {\includegraphics[width=0.115\textwidth]{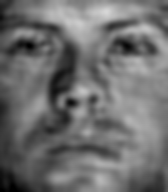}}
      \subcaptionbox{\label{subfig:e}} {\includegraphics[width=0.115\textwidth]{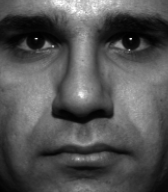}}
    \subcaptionbox{ $\sigma_Z^2 = 0.15$ \label{subfig:f}} {\includegraphics[width=0.115\textwidth]{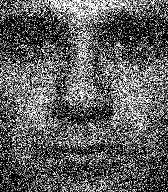}}
    \subcaptionbox{$25.82$ dB\label{subfig:g}} {\includegraphics[width=0.115\textwidth]{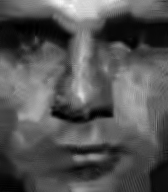}}
   \subcaptionbox{$25.77$ dB\label{subfig:h}} {\includegraphics[width=0.115\textwidth]{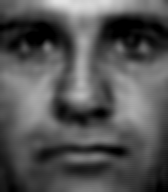}}

   \end{center}
   \caption{Samples of image denoising. Order of columns: original image, noisy (noise variance), BM3D (PSNR) and RRQ (PSNR). }
   \label{fig:Images_denoising}
   \end{figure}

Fig. \ref{fig:Images_denoising} illustrates the denoising quality for two image samples. It is interesting to notice that the BM3D, although producing a smooth image, fails to reconstruct the face contours since it lacks enough priors.

%

\section{Conclusions}
\label{summary}
A framework for multi-layer representation of images was proposed where, instead of local patch-based processing, a global high-dimensional vector representation of images is successively quantized within different levels of reconstruction fidelity. As an alternative to the classical RQ framework which is based on k-means, the proposed RRQ along with pre-processing, randomly generates codewords from a regularized and learned distribution. Apart from the many potential advantages of having random codewords, this is shown lo lead to efficient quantization with low train-test distortion gaps. The experimental results show interesting promise for different practical scenarios, e.g., when the acquisition devices at the query phase are much noisier than the enrollment cameras. Future works consider using the variance priors to further train the codewords, moreover using entropy coding on the tree of indices for better rate-distortion performance. 
\clearpage
\bibliographystyle{IEEEbib}

\bibliography{tempBib}

\end{document}